\documentclass[11pt]{article}

\usepackage[preprint]{acl}

\usepackage{times}
\usepackage{latexsym}
\usepackage[table]{xcolor}
\usepackage[T1]{fontenc}
\usepackage[utf8]{inputenc}

 \usepackage{amsmath}
\usepackage{microtype}
\usepackage{microtype}
\usepackage{graphicx}
\usepackage{float} % Optional: helps with placement
\usepackage{subfigure}
\usepackage{booktabs} % for professional tables
\usepackage{tcolorbox}
\usepackage{hyperref}

\usepackage{booktabs}
\usepackage{multirow}
\usepackage{adjustbox}
\usepackage{colortbl} % Optional, for gray rows if needed
\definecolor{lightgray}{gray}{0.9}
\usepackage{inconsolata}
\usepackage{algorithm}
\usepackage{algorithmic}
\usepackage{graphicx}

\title{Small Reasoning Models are Instruction Followers in Function Calling}

\author{
  Yalda Taheri \\
  Department of Engineering \\
  Islamic Azad University \\
  \texttt{y.jaliltahei@iau.ir} \\
  \And
  Mohammad Hassan Heydari \\
  Department of Computer Engineering \\
  University of Isfahan \\
  \texttt{m.heydari@mehr.ui.ac.ir} \\
  \AND
  Erfan Naaman \\
  Department of Computer Engineering \\
  University of Isfahan \\
  \texttt{erfannamaan@mehr.ui.ac.ir} \\
  \And
  Afsaneh Fatemi \\
  Department of Computer Engineering \\
  University of Isfahan \\
  \texttt{a\_fatemi@eng.ui.ac.ir}
}

\begin{document}
\maketitle
\begin{abstract}
Function calling represents the core capability of agentic large language models (LLMs). Existing research has focused on enhancing LLMs’ function-calling accuracy through fine-tuning, reinforcement learning (RL), and multi-agent frameworks, particularly for native function-calling LLMs. This work demonstrates that LLMs achieve superior accuracy in function calling in instruction-following contexts (i.e., standard user-assistant interactions) rather than a tool calling context. We introduce Instruction-Followed Function Calling (IFFC), a novel framework that decouples function-calling logic from the primary LLM and delegates it to a dedicated smaller model operating within the instruction-following paradigm. Our method consistently outperforms both native function calling (NFC) and prompt-based function calling (PFC) baselines, with particularly strong gains on reasoning-oriented LLMs. Furthermore, we demonstrate that IFFC maintains robust performance under aggressive quantization, enabling efficient on-device deployment without significant accuracy degradation. This work establishes a new paradigm for reliable, resource-efficient function calling in edge-computing scenarios.
\end{abstract}

\section{Introduction}
The evolution of artificial intelligence has transitioned from static text generation to autonomous, active problem-solving, a paradigm known as "Agentic AI" \cite{bfcl}. At the core of this transition is function calling, which enables models to interact with external environments by selecting tools, generating arguments, and executing actions \cite{kavathekar2025small}. While massive proprietary models initially dominated this space, there is a growing shift toward specialized, task-specific agents that are often more suitable for structured and repetitive workflows than generalist architectures \cite{belcak2025small, zeng2025routine}.
Simultaneously, Small Language Models (SLMs) ranging from 0.5 to 15 billion parameters have emerged as a practical alternative to massive models, which often suffer from high computational demands, latency, and privacy risks \cite{samoylenko2025position, xu2024small}. When specialized, these compact models can rival their larger counterparts in domain-specific applications, such as fault diagnosis and code generation, while remaining accessible to edge devices and common hardware \cite{kumar2025building, nath2025domain, sinha2025small}. This makes the intersection of SLMs and localized agentic workflows a highly promising area of deployment.
However, enabling robust function calling within SLMs presents distinct challenges, as these smaller models frequently struggle with rigid syntactic constraints, complex JSON schemas, and multi-step reasoning \cite{kavathekar2025small, sharma2025small}. Although techniques such as targeted fine-tuning, reinforcement learning, and advanced prompt engineering have been proposed to mitigate these limitations \cite{jhandi2025small, paprunia2025advancing, han2025enhancing}, many of these approaches continue to force SLMs into native function-calling paradigms designed for larger models. Consequently, this often leads to suboptimal format adherence and fragile execution logic in constrained environments.

\begin{figure*}[t]  % Added the asterisk (*) here
    \centering
    \includegraphics[width=0.9\textwidth]{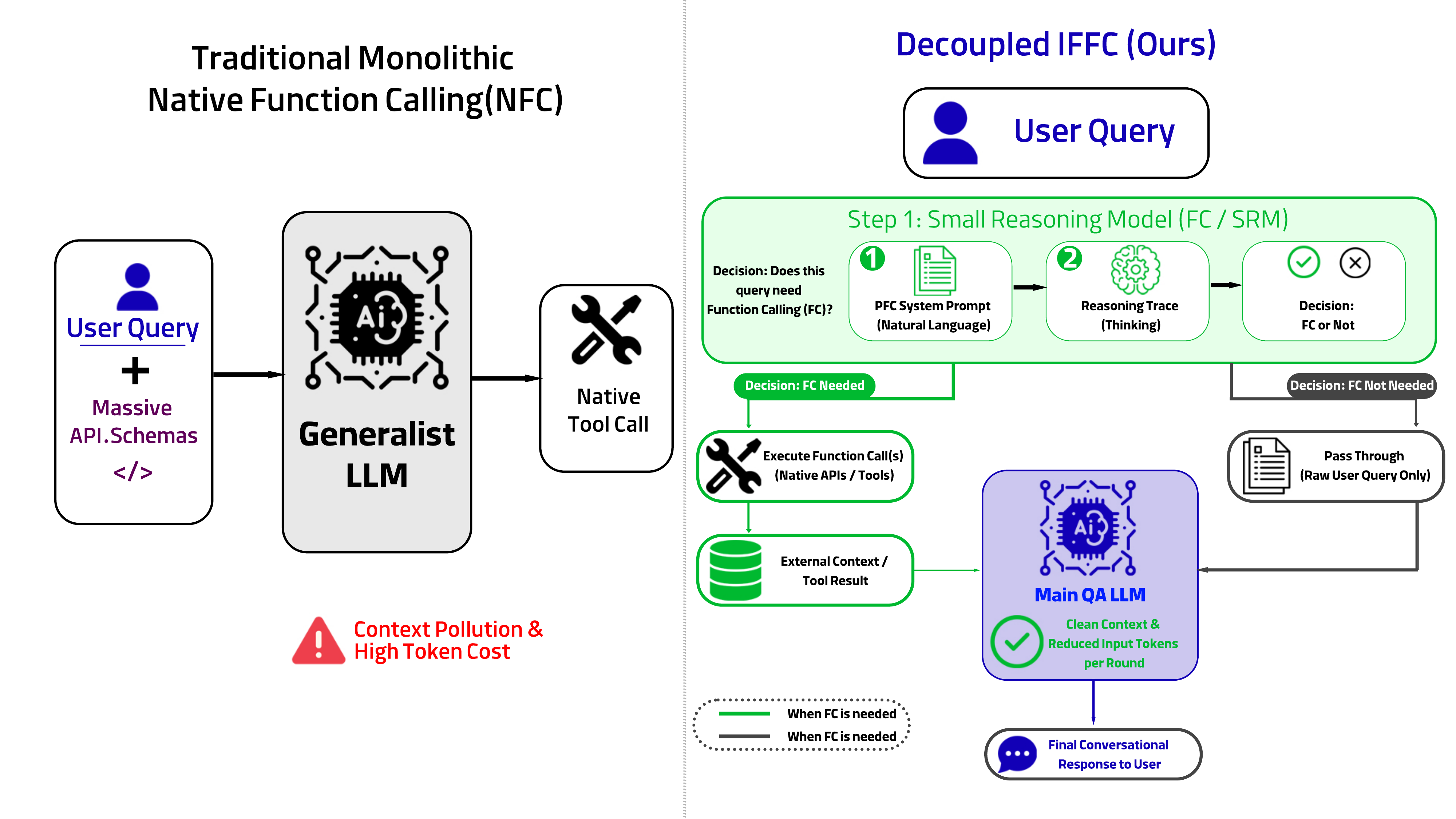} % Adjust width as needed
    \caption{Overview of Instruction-Followed Function Calling (IFFC) framework against traditional function calling approach.}
    \label{fig:figure22}
\end{figure*}       % Added the asterisk (*) here

In this paper, we challenge the prevailing reliance on native function calling (NFC) for agentic tasks. We posit that the architectural strengths of language models, particularly those optimized for reasoning, lie in following natural language instructions rather than manipulating rigid tool definitions \cite{johnson2025natural}. We introduce Instruction-Followed Function Calling (IFFC), a framework that re-imagines the tool-use process by decoupling the function calling task from the main question-answering (QA) model and assign it to another model (in our case, an SLM) which separates their contexts entirely \cite{roth2025factored, jeon2025slm}. By delegating the function-calling mechanism to a dedicated, instruction-following paradigm with Reason-Action (ReAct) mechanism, we bypass the limitations inherent in native tool support.
Our contributions are as follows:
\begin{itemize}
\item We demonstrate that, contrary to common practice, models achieve superior function-calling accuracy in instruction-following contexts (a.k.a. regular user-assistant prompting context) compared to native tool-calling contexts.
\item We propose the IFFC framework, which first uses PFC as its tool calling method and then enables smaller reasoning models (SRMs) to execute complex tool usage with higher reliability than native baselines, while being completely separated from the main QA LLM. We show that this context separation allows the SRM to be less context polluted by the outputs of the QA LLM. This hypothesis, which assumes that reasoning models perform better than non-reasoning models, is validated throughout our experiments
\item We validate the efficiency of our approach by showing that IFFC maintains robust performance even under aggressive quantization. This establishes a viable path for deploying accurate, privacy-preserving agentic capabilities on resource-constrained edge devices.
\end{itemize}

\section{Related Works}
Function calling has transitioned language models from passive text generators to active agents capable of automating complex tasks \cite{kavathekar2025small}. While early agentic frameworks relied on massive, general-purpose models, recent research emphasizes that small language models (SLMs) under 7 billion parameters offer a more stable, cost-effective, and private alternative for structured workflows \cite{sharma2025small, samoylenko2025position}. When properly specialized, these compact architectures can rival larger systems on targeted reasoning tasks, bypassing the significant computational and financial overhead associated with proprietary models \cite{sinha2025small, xu2024small}.
Deploying SLMs for function calling introduces unique architectural opportunities, such as guided decoding, type-safe registries, and decoupled structures that separate planning from execution to minimize formatting errors \cite{sharma2025small, roth2025factored}. These optimizations enable efficient local deployment on resource-constrained edge devices, as demonstrated by frameworks like TinyAgent \cite{erdogan2024tinyagent}. Furthermore, techniques like quantization and domain-specific adaptation allow SLMs to excel in low-latency and privacy-sensitive applications, including automotive control, smart home assistance, and specialized domain tasks \cite{khiabani2025optimizing, huang2026towards, jia2025beyond, nath2025domain}.
To narrow the performance gap between SLMs and larger models, researchers have investigated both training-based and training-free optimization strategies. Training-based methods leverage reinforcement learning, group relative policy optimization (GRPO), domain-specific fine-tuning, and planning distillation from larger teacher agents to reinforce structured tool-use behavior \cite{qian2025toolrl, paprunia2025advancing, jhandi2025small, qiu2025agentdistill}. Alternatively, training-free approaches utilize natural language tool interfaces, advanced prompt engineering, reasoning blueprints, and iterative trial-and-error frameworks to improve tool selection and execution without additional training overhead \cite{johnson2025natural, he2024achieving, han2025enhancing, qu2024exploration}.

\begin{table*}[t]
\centering
\caption{Impact of Q4KM Quantization on Memory Usage and IFFC Accuracy (BFCL V3 Non-Live)}
\label{tab:quantization_impact}
\small
\begin{tabular}{@{} l ccc c cccc @{}}
\toprule
& \multicolumn{3}{c}{\textbf{Memory Footprint}} & \phantom{a} & \multicolumn{4}{c}{\textbf{Accuracy Degradation}} \\
\cmidrule(lr){2-4} \cmidrule(l){6-9}
\textbf{Model} & \textbf{FP16} & \textbf{Q4KM} & \textbf{Reduction} && \textbf{FP16} & \textbf{Q4KM} & \textbf{$\Delta$ (pts)} & \textbf{Rel. Drop} \\
\midrule
Gemma-3 1B            & 2.0 GB  & 0.8 GB & 60.0\% && 23.9\% & 8.7\%  & --15.2 & 63.6\% \\
Gemma-3 4B            & 8.6 GB  & 3.3 GB & 61.6\% && 81.9\% & 60.4\% & --21.5 & 26.3\% \\
Gemma-3 12B           & 24.3 GB & 8.1 GB & 66.7\% && 91.3\% & 69.0\% & --22.3 & 24.4\% \\
Phi-4 Mini            & 7.7 GB  & 2.5 GB & 67.5\% && 74.9\% & 48.2\% & --26.7 & 35.6\% \\
\midrule
Qwen-3 0.6B (NoThink) & 1.5 GB  & 0.5 GB & 66.7\% && 62.3\% & 16.8\% & --45.5 & 73.0\% \\
Qwen-3 0.6B (Think)   & 1.5 GB  & 0.5 GB & 66.7\% && 76.4\% & 69.2\% & --7.2  & 9.4\%  \\
Qwen-3 1.7B (NoThink) & 4.0 GB  & 1.4 GB & 65.0\% && 82.8\% & 80.5\% & --2.3  & 2.8\%  \\
Qwen-3 1.7B (Think)   & 4.0 GB  & 1.4 GB & 65.0\% && 82.8\% & 82.9\% & +0.1   & --0.1\%\\
Qwen-3 4B (NoThink)   & 8.0 GB  & 2.5 GB & 68.8\% && 91.9\% & 91.0\% & --0.9  & 1.0\%  \\
Qwen-3 4B (Think)     & 8.0 GB  & 2.5 GB & 68.8\% && 94.1\% & 93.5\% & --0.6  & 0.6\%  \\
Qwen-3 8B (NoThink)   & 16.4 GB & 5.2 GB & 68.3\% && 93.8\% & 92.8\% & --1.0  & 1.1\%  \\
Qwen-3 8B (Think)     & 16.4 GB & 5.2 GB & 68.3\% && 94.2\% & 94.4\% & +0.2   & --0.2\%\\
\bottomrule
\end{tabular}
\end{table*}

\section{Methodology}

\begin{table}[htbp]
\centering
\caption{Average Function Calling Accuracy (\%) Over Four Categories on BFCL V3}
\label{tab:average_accuracy_results}
\small
\begin{tabular}{l >{\columncolor{gray!15}}c c c} % <-- Applied column color here
\toprule
\textbf{Model} & \textbf{IFFC (Ours)} & \textbf{PFC} & \textbf{NFC} \\
\midrule
\multicolumn{4}{l}{\textbf{BFCL V3 Live (Average)}} \\
Phi-4 Mini                  & 44.3          & \textbf{62.5} & 30.0 \\
Qwen-3 0.6B Think           & \textbf{57.7} & 53.2          & 53.0 \\
Qwen-3 1.7B Think           & 72.8          & 73.6          & \textbf{74.6} \\
Qwen-3 4B Think             & \textbf{86.7} & 82.8          & 81.5 \\
Qwen-3 8B Think             & \textbf{83.3} & 78.3          & 75.5 \\
\midrule
\multicolumn{4}{l}{\textbf{BFCL V3 Non-Live (Average)}} \\
Phi-4 Mini                  & \textbf{77.4} & 41.9          & 9.5  \\
Qwen-3 0.6B Think           & \textbf{76.5} & 72.8          & 71.8 \\
Qwen-3 1.7B Think           & \textbf{84.0} & 65.3          & 83.0 \\
Qwen-3 4B Think             & \textbf{94.1} & 88.7          & 88.6 \\
Qwen-3 8B Think             & \textbf{94.9} & 89.7          & 88.8 \\
\bottomrule
\end{tabular}
\end{table}

We introduce a two-stage paradigm: first, decoupling the tool-selection logic from the primary generation model, and second, re-framing the function-calling task as a standard instruction-following interaction.

\subsection{Decoupling Function Calling from the Main LLM}

Traditional agentic workflows often utilize a single monolithic LLM to handle both the reasoning required for tool selection and the final synthesis of the response. This approach frequently leads to context pollution, where the presence of complex API schemas in the prompt interferes with the model's conversational performance or reasoning depth \cite{roth2025factored, bfcl}. 

Drawing inspiration from the success of multi-agent systems in isolating specific sub-tasks \cite{zeng2025routine}, we propose a decoupled agentic framework as illustrated in Figure \ref{fig:figure22} and Algorithm \ref{alg:iffc_workflow}. We introduce a dedicated \textbf{SRM}, typically ranging from 0.5B to 15B parameters, to serve as the primary routing and tool-execution layer. When an \textit{Incoming Query} is received, the SRM evaluates the user's intent to determine if external context or tool execution is required to provide an accurate answer. 

By delegating this logic to a smaller, faster, and more cost-effective model, we achieve several advantages:
\begin{itemize}
    \item \textbf{Context Isolation:} The main LLM's context remains focused on the user interaction and the final response generation, while the SRM handles the "heavy lifting" of tool-definition parsing \cite{belcak2025small}.
    \item \textbf{Efficiency:} The SRM can be aggressively quantized and deployed on the edge, significantly reducing latency compared to routing every query through a massive generalist LLM \cite{erdogan2024tinyagent}.
    \item \textbf{Token Optimization per Turn:} Because the SRM's context is completely isolated from the main LLM's conversational synthesis, the final generated response (which can be several hundred tokens long) is never appended to the SRM's conversational history. This limits the growth of the input context window for the SRM in multi-turn interactions.
\end{itemize}

To formalize the decoupled paradigm, Algorithm \ref{alg:iffc_workflow} in Appendix \ref{sec:pfc} details the step-by-step execution flow of the IFFC framework. When a query is initiated, the SRM evaluates the intent using PFC instead of NFC. If tool execution is deemed necessary, the system executes the function and appends the resulting context directly to the query. 

\subsection{Instruction-Followed Function Calling (IFFC)}

Our second major finding is that models, especially SRMs, demonstrate higher accuracy when performing tool selection within a standard "Instruction-Following" context rather than a specialized "Tool Calling" context. Most modern LLMs are trained with native function-calling (NFC) support, requiring specific tags (e.g., \texttt{<tool\_declare>} and \texttt{<tool\_call>}) and rigid JSON schemas. However, empirical results from the Berkeley Function Calling Leaderboard (BFCL) \cite{bfcl} and our own experiments suggest that these rigid constraints often lead to formatting errors and hallucinations in smaller models \cite{kavathekar2025small, johnson2025natural}; Which eventually shows that LLMs are more capable of function calling when they are prompted, rather than when they are declared in special tool tokens. 

We introduce \textbf{Instruction-Followed Function Calling (IFFC)}, which bypasses the native API-call wrappers in favor of natural language instructions. As shown in the comparison in our framework's prompt structure:
\begin{itemize}
    \item \textbf{Native Function Calling (NFC):} Uses specialized, non-conversational tokens to declare and call tools, which can be brittle and sensitive to formatting errors.
    \item \textbf{IFFC Paradigm:} Treats the tool definition as a high-priority system instruction and the function call as a standard assistant response. 
\end{itemize}

By framing the task as a regular user-assistant interaction, we leverage the extensive instruction-tuning that these models undergo. Instead of the SRM struggling with the syntactic overhead of NFC, it follows a system prompt that explicitly defines the tools as part of its "behavioral guidelines." This transition from \textit{manipulating rigid definitions} to \textit{following conversational instructions} allows the SRM to focus its reasoning capacity on argument generation and tool selection logic, leading to the performance gains observed in Tables \ref{tab:average_accuracy_results}, \ref{tab:bfcl_results} and \ref{tab:gemma3_iffc_vs_pfc}.

\begin{table*}[t]
\centering
\caption{Model Accuracy (\%) Comparison on BFCL V3 Live and Non-Live Benchmarks}
\label{tab:bfcl_results}
\footnotesize % Changed from \small to \footnotesize to fit the margins
\setlength{\tabcolsep}{3.5pt} % Safely reduces horizontal padding between columns
\begin{tabular}{@{} l cccc c cccc @{}}
\toprule
& \multicolumn{4}{c}{\textbf{BFCL V3 Live}} & \phantom{a} & \multicolumn{4}{c}{\textbf{BFCL V3 Non-Live}} \\
% Trimmed both midrules (lr) to ensure a clean visual gap in the middle
\cmidrule(lr){2-5} \cmidrule(lr){7-10}
\textbf{Model} & \textbf{Simple} & \textbf{Multiple} & \textbf{Parallel} & \textbf{Par. Mult.} && \textbf{Simple} & \textbf{Multiple} & \textbf{Parallel} & \textbf{Par. Mult.} \\
\midrule
\rowcolor{gray!10} Qwen-3 4B Think IFFC (Ours) & \textbf{90.3} & \textbf{81.5} & \textbf{87.5} & \textbf{87.5} && \textbf{96.0} & \textbf{97.5} & 92.5 & 90.5 \\
Qwen-3 4B Think NFC        & 87.6 & 79.9 & 75.0 & 83.3          && 75.3 & 96.5 & 92.0 & 90.5 \\
Claude 4.5 Sonnet NFC      & 89.5 & 78.9 & \textbf{87.5} & 83.3          && 72.6 & 95.5 & \textbf{94.5} & \textbf{92.0} \\
GPT 5.2 NFC                & 71.7 & 70.4 & 68.8 & 58.3          && 72.9 & 88.0 & 89.0 & 77.5 \\
Gemini 2.5 Pro NFC         & 77.9 & 62.2 & 68.8 & 62.5          && 66.4 & 86.0 & 69.0 & 40.0 \\
\bottomrule
\end{tabular}
\end{table*}

\section{Experiments}

\label{sec:experiments}

To validate the efficacy of the IFFC framework, we conducted a comprehensive empirical analysis comparing it against established NFC and PFC paradigms. Our experimental design focuses on three key dimensions: the comparative performance of SLMs, specially SRMs, against state-of-the-art proprietary models, the impact of "thinking" modes in hybrid reasoning architectures, and the robustness of our approach under quantization for edge deployment.

We benchmarked our method using a diverse suite of open-weights models to represent the landscape of efficient SRMs. Specifically, we evaluated the \textbf{Gemma-3} series (1B, 4B, and 12B) \cite{gemma3}, \textbf{Phi-4 Mini Instruct} \cite{phi4mini}, and the \textbf{Qwen-3} series (0.6B, 1.7B, 4B, and 8B) \cite{qwen3}. Additionally, to test the limits of extreme compression, we included \textbf{Granite 4 Micro} and \textbf{Granite 4 Tiny-h} \cite{granite} (detailed specifications provided in the Appendix \ref{sec:full_results}).

To establish a rigorous baseline, we compared these SRMs operating under IFFC against the current state-of-the-art proprietary models operating in their native function calling (NFC) modes. These baselines include \textbf{GPT 5.2}, \textbf{Gemini 2.5 Pro} \cite{gemini}, and \textbf{Claude 4.5 Sonnet}. This comparison aims to determine if decoupled SRMs can rival the performance of massive generalist models in tool-use scenarios.

To isolate the effect of reasoning, we utilized the \textbf{Qwen-3} \cite{qwen3} hybrid models, which support toggleable inference modes. This ablation study allows us to quantify how much the explicit reasoning trace contributes to accurate argument parsing and schema adherence compared to standard generation.

For IFFC to be a viable "plug-and-play" solution for edge AI, it must maintain performance when models are compressed. We conducted a sensitivity analysis by comparing the performance of our selected models at \textbf{FP16} versus \textbf{Q4KM} quantization levels. 

Given that IFFC shares architectural similarities with PFC, in that both rely on natural language prompts rather than specialized tokens, we performed a direct comparison between the two methods using the \textbf{Gemma-3} \cite{gemma3} model family. Full details of the differences between IFFC and PFC is discussed in Appendix \ref{sec:pfc}

\section{Results}
\label{sec:results}

\subsection{Superiority of Instruction Following over Native Tool Use}
\label{subsec:results_main}

Our primary hypothesis was that SRMs perform better when tool execution is framed as a conversational instruction rather than a rigid schema constraint. The results presented in Tables \ref{tab:bfcl_results} and \ref{tab:average_accuracy_results} strongly corroborate this. 

Across the Qwen-3 series (0.6B to 8B) and Phi-4 Mini, IFFC consistently outperforms the NFC baseline. The disparity is particularly pronounced in smaller models; for instance, Phi-4 Mini achieves only $\sim$30\% accuracy in NFC mode (Live) but jumps to \textbf{44.3\%} using IFFC. Similarly, the Qwen-3 4B (Think) model sees a substantial improvement, reaching \textbf{86.7\%} in Live evaluation and \textbf{94.1\%} in Non-Live evaluation, significantly surpassing both the PFC (82\%) and NFC (81\%) baselines.

We further analyzed the limitations of PFC using the Gemma-3 family. As shown in Table \ref{tab:gemma3_iffc_vs_pfc}, while PFC performs adequately on "Simple" queries, it suffers catastrophic degradation in complex scenarios. In the "Parallel Multiple" category, Gemma-3 4B using PFC drops to near 0\% accuracy due to context drift and hallucination. In contrast, the IFFC framework maintains robustness, with the Gemma-3 12B model achieving \textbf{79.1\%} accuracy in the same category. This confirms that decoupling the routing logic allows models to handle high-complexity queries without the context pollution inherent in standard prompting methods.

\begin{table*}[t]
\centering
\caption{Comparison of Gemma-3 Models using IFFC vs. PFC across BFCL V3 Live and Non-Live Categories.}
\label{tab:gemma3_iffc_vs_pfc}
\footnotesize 
\setlength{\tabcolsep}{8pt} % Slightly relaxed column padding since we removed a column
\begin{tabular}{@{} lcccc @{}} % Reduced to 5 columns and removed outer margins
\toprule
\textbf{Model \& Method} & \textbf{Simple (\%)} & \textbf{Multiple (\%)} & \textbf{Parallel (\%)} & \textbf{Par. Mult. (\%)} \\
\midrule

% Section Header 1
\multicolumn{5}{@{} l}{\textbf{BFCL V3 Live}} \\
\midrule
\rowcolor{gray!10} Gemma-3 1B IFFC (Ours) & 13.9 & 7.1  & \textbf{31.3} & \textbf{8.3}  \\
Gemma-3 1B PFC         & \textbf{30.0} & \textbf{10.5} & 0.0  & 0.0  \\
\cmidrule{1-5}
\rowcolor{gray!10} Gemma-3 4B IFFC (Ours) & \textbf{77.1} & \textbf{63.7} & \textbf{75.0} & \textbf{54.2} \\
Gemma-3 4B PFC         & 72.9 & 62.8 & 37.5 & 29.2 \\
\cmidrule{1-5}
\rowcolor{gray!10} Gemma-3 12B IFFC (Ours)& \textbf{86.4} & \textbf{78.5} & \textbf{87.5} & \textbf{79.2} \\
Gemma-3 12B PFC        & 84.9 & 70.9 & \textbf{87.5} & 62.5 \\
\midrule

% Section Header 2
\multicolumn{5}{@{} l}{\textbf{BFCL V3 Non-Live}} \\
\midrule
\rowcolor{gray!10} Gemma-3 1B IFFC (Ours) & 21.8 & 36.0 & \textbf{22.0} & \textbf{16.0} \\
Gemma-3 1B PFC         & \textbf{43.5} & \textbf{38.5} & 2.0  & 2.0  \\
\cmidrule{1-5}
\rowcolor{gray!10} Gemma-3 4B IFFC (Ours) & \textbf{87.6} & 85.0 & \textbf{82.5} & \textbf{72.5} \\
Gemma-3 4B PFC         & 64.3 & \textbf{91.5} & 56.5 & 41.0 \\
\cmidrule{1-5}
\rowcolor{gray!10} Gemma-3 12B IFFC (Ours)& \textbf{94.0} & 93.5 & \textbf{90.0} & \textbf{89.0} \\
Gemma-3 12B PFC        & 77.3 & \textbf{95.0} & \textbf{90.0} & 73.0 \\
\bottomrule
\end{tabular}
\end{table*}

\subsection{Small Reasoning Models vs. Proprietary Giants}
\label{subsec:results_sota}

The results, illustrated in Table \ref{tab:bfcl_results}, demonstrate that our decoupled SRM approach is highly competitive. In the "Non-Live" evaluation, Qwen-3 4B (IFFC) achieves \textbf{96.0\%} on Simple tasks and \textbf{97.5\%} on Multiple tasks, outperforming Claude 4.5 Sonnet (72.6\% and 95.5\% respectively) and GPT-5.2 (72.9\% and 88.0\% respectively). Even in the challenging "Parallel Multiple" category, Qwen-3 4B achieves \textbf{90.5\%}, which is comparable to Claude 4.5 Sonnet (92.0\%) and significantly higher than Gemini 2.5 Pro (40.0\%). This indicates that a specialized 4B parameter model, when relieved of the syntactic burden of native API definitions, can match or exceed the reasoning fidelity of models orders of magnitude larger.

\begin{table}[htbp]
\centering
\caption{Performance Comparison of Qwen-3 in No-Think vs. Think Modes under the IFFC Framework}
\label{tab:qwen3_think_comparison}
\resizebox{\columnwidth}{!}{% <-- Scales the table horizontally to fit the column width perfectly
\begin{tabular}{lccccc}
\toprule
 & \multicolumn{2}{c}{\textbf{BFCL V3 Live (\%)}} & & \multicolumn{2}{c}{\textbf{BFCL V3 Non-Live (\%)}} \\
\cmidrule{2-3} \cmidrule{5-6}
\textbf{Model} & \textbf{No-Think} & \textbf{Think} & & \textbf{No-Think} & \textbf{Think} \\
\midrule
Qwen-3 0.6B & 22.9 & \textbf{57.7} & & 62.3 & \textbf{76.5} \\
Qwen-3 1.7B & 65.0 & \textbf{72.8} & & 81.1 & \textbf{84.0} \\
Qwen-3 4B   & 74.6 & \textbf{86.7} & & 91.9 & \textbf{94.1} \\
Qwen-3 8B   & 74.5 & \textbf{83.3} & & 93.8 & \textbf{94.9} \\
\bottomrule
\end{tabular}%
}
\end{table}

\subsection{The Impact of Reasoning ("Thinking") on Tool Selection}
\label{subsec:results_thinking}

Table \ref{tab:qwen3_think_comparison} presents the comparison between "Think" (Reasoning enabled) and "No-Think" (Standard generation) modes under the IFFC framework.

The reasoning effect is most dramatic in the smallest models; Qwen-3 0.6B sees its accuracy arguably double, jumping from 22.9\% to \textbf{57.7\%} in Live evaluation when reasoning is enabled. For the 4B model, the "Think" mode pushes accuracy from 74.6\% to \textbf{86.7\%}.

\subsection{Robustness to Quantization}
\label{subsec:results_quant}

The results in Table \ref{tab:quantization_impact}  reveal a critical divergence between standard instruction models and reasoning models.

Standard models suffer significant degradation under quantization; for example, Gemma-3 1B drops from 23.9\% to \textbf{8.7\%}, and the standard Qwen-3 0.6B (No-Think) drops from 62.3\% to \textbf{16.8\%}. However, reasoning-oriented models exhibit remarkable resilience. The Qwen-3 4B (Think) model maintains \textbf{93.5\%} accuracy at Q4KM, a negligible drop from 94.1\% at FP16. Similarly, the 8B (Think) model actually shows a slight variance improvement to 94.4\%.

This finding suggests that the "reasoning trace" acts as a form of error correction that compensates for the precision loss in model weights, making SRMs uniquely working for efficient on-device function calling. Full experiment results are presented in Appendix \ref{sec:full_results}

\section{Conclusion}
\label{sec:conclusion}

In this work, we introduced Instruction-Followed Function Calling (IFFC), a framework that fundamentally redefines how Small Reasoning Models (SRMs) execute agentic tasks by prioritizing natural language instruction adherence over rigid native tool definitions. Our extensive empirical evaluation demonstrates that decoupling the routing logic enables compact models, such as the Qwen-3 4B, to outperform massive proprietary baselines like GPT-5.2 and Claude 4.5 Sonnet, particularly when leveraging explicit reasoning traces. Furthermore, we validated the remarkable robustness of reasoning-oriented models under aggressive quantization, confirming their viability for efficient, privacy-preserving edge deployment.

\section{Limitations}
\label{sec:limitations}
While the Instruction-Followed Function Calling (IFFC) framework demonstrates significant accuracy improvements and robustness under quantization, the decoupled two-stage architecture introduces inherent latency and scalability trade-offs. By delegating intent evaluation and tool routing to a dedicated Small Reasoning Model (SRM) before passing the enriched context to the primary QA model, the system necessitates sequential inference steps. Although aggressively quantizing the SRM to Q4KM mitigates memory footprint and computational overhead, real-time applications with strict latency budgets may still experience delays.

% \bibliography{example_paper}
% \bibliographystyle{icml2025}

%%%%%%%%%%%%%%%%%%%%%%%%%%%%%%%%%%%%%%%%%%%%%%%%%%%%%%%%%%%%%%%%%%%%%%%%%%%%%%%
%%%%%%%%%%%%%%%%%%%%%%%%%%%%%%%%%%%%%%%%%%%%%%%%%%%%%%%%%%%%%%%%%%%%%%%%%%%%%%%
% APPENDIX
%%%%%%%%%%%%%%%%%%%%%%%%%%%%%%%%%%%%%%%%%%%%%%%%%%%%%%%%%%%%%%%%%%%%%%%%%%%%%%%
%%%%%%%%%%%%%%%%%%%%%%%%%%%%%%%%%%%%%%%%%%%%%%%%%%%%%%%%%%%%%%%%%%%%%%%%%%%%%%%

\bibliography{custom}

\newpage
\appendix

\onecolumn

\section{IFFC and PFC: A review on differences}
\label{sec:pfc}
\subsection{The Necessity of Prompting}
Not all Large Language Models (LLMs) are trained with native function-calling capabilities. Several high-performance open-weights models, such as the \textbf{Gemma-3} \cite{gemma3} family, do not utilize specialized control tokens (e.g., \texttt{<tool\_call>}) or separate API-calling heads. To enable function calling in these architectures, we utilize Prompt-based Function Calling (PFC). 

In this paradigm, we inject a robust system prompt that defines the model's persona as an expert in function execution. The available tools are serialized into the system context, and the model is instructed to invoke them when necessary. Crucially, unlike Native Function Calling (NFC) which outputs to a specialized \texttt{tool\_use} role, PFC models generate the function invocation directly within the standard \texttt{assistant} response role.

\subsection{Relation to IFFC}
Our Instruction-Followed Function Calling (IFFC) framework shares the fundamental nature of PFC: it treats tool usage as a text-generation task governed by strict instruction adherence. However, IFFC diverges from standard PFC baselines (such as those found in the BFCL repository) in two key ways:

\begin{enumerate}
    \item \textbf{Decoupled Architecture:} Standard PFC usually feeds the tool definitions to the main conversational model. IFFC offloads this entirely to a dedicated Small Reasoning Model (SRM), isolating the context.
    \item \textbf{Customized Formatting:} While standard benchmarks often demand a generic JSON format, IFFC enforces a specialized output format optimized for the SRM's reasoning capabilities.
\end{enumerate}

\subsection{Targeted Prompt Engineering and Heuristics}
Through iterative testing on the Berkeley Function Calling Leaderboard (BFCL) \cite{bfcl}, we identified specific weaknesses where even capable models struggled with rigid evaluation criteria. To address this, we integrated targeted hints into the IFFC system prompt to guide the SRM's logic.

A primary example involves parameter formatting. We observed that models often hallucinated argument formats that were semantically correct but syntactically mismatched with the API definition (e.g., providing a city name alone when the API required a specific string pattern). We explicitly updated the system instruction to force adherence to examples provided in the function docstrings. 

For instance, for a function \texttt{get\_weather(location: str)}, if the documentation provides an example value like \texttt{"City, Country"}, our prompt explicitly instructs the model to: \textit{"Use the exact same format of function parameter values as the examples in the function definition."} This ensures that the model generates \texttt{"Paris, France"} rather than just \texttt{"Paris"}, significantly reducing schema validation errors during evaluation.

\section{IFFC System Prompt}
To ensure reproducibility, we provide the exact system prompt utilized in the Instruction-Followed Function Calling (IFFC) framework. The \texttt{\{available\_functions\}} placeholder is dynamically populated with the specific tool definitions relevant to the query.

\begin{tcolorbox}[colback=gray!10, colframe=gray!60, arc=2mm, title=IFFC System Instruction]
\small\texttt{
You are an expert in function calling. You will be given a set of available functions and a user query.
You should determine if the user query needs function calling, call the appropriate function with suitable arguments.
Based on the question , you will need to make one or more function calls to achieve the purpose.
\newline
\newline
Here are the available functions:
\\\{available\_functions\}
\newline
\newline
For every user query you only answer in json format. You will return a list of dictionaries. If the user question doesn't require any function calling you will return an empty list ( e.g. [] ).
If it does require function calling, for every function you add a dictionary which has two keys, "function\_name" and "parameters".
"parameters" is also a dict with function arguments and their values.
Use the exact same format of function parameter values as the examples in the function definition.
}
\end{tcolorbox}

\subsection{Decoupling in IFFC}

One of the core differences between traditional function calling and IFFC is the context separation of the QA LLM and the function caller, which leads the function caller to be less context polluted. In Algorithm \ref{alg:iffc_workflow} we provide a detailed schema of how context separation operates in IFFC.

\begin{algorithm*}[t]
\caption{Instruction-Followed Function Calling (IFFC) Workflow}
\label{alg:iffc_workflow}
\begin{algorithmic}[1]
\REQUIRE User Query $Q$, Available Functions $\mathcal{F}$, SRM System Prompt $P_{\text{SRM}}$ (with IFFC guidelines), Main LLM System Prompt $P_{\text{Main}}$, SRM Conversational State $\mathcal{H}_{\text{SRM}}$, Main LLM Conversational State $\mathcal{H}_{\text{Main}}$
\ENSURE Final Conversational Response $R$
\STATE 
\STATE // \textit{Stage 1: Intent evaluation and function routing via SRM}
\STATE Construct input payload for SRM: $X_{\text{SRM}} \leftarrow \text{FormatPrompt}(P_{\text{SRM}}, \mathcal{F}, Q, \mathcal{H}_{\text{SRM}})$
\STATE Generate SRM response: $O_{\text{SRM}} \leftarrow \text{SRM}(X_{\text{SRM}})$ \COMMENT{Executed via Prompt-based Function Calling (PFC)}
\STATE
\STATE // \textit{Analyze output for required external knowledge}
\IF{$O_{\text{SRM}}$ triggers a function call $f \in \mathcal{F}$ with parameters $\theta$}
    \STATE Execute external function: $C \leftarrow \text{Execute}(f, \theta)$ \COMMENT{Retrieve external context}
    \STATE Formulate context-enriched query: $Q_{\text{enriched}} \leftarrow \text{Combine}(Q, C)$
\ELSE
    \STATE $Q_{\text{enriched}} \leftarrow Q$
\ENDIF
\STATE
\STATE // \textit{Stage 2: Independent Response Synthesis via Main QA LLM}
\STATE Construct input payload for Main LLM: $X_{\text{Main}} \leftarrow \text{FormatPrompt}(P_{\text{Main}}, Q_{\text{enriched}}, \mathcal{H}_{\text{Main}})$
\STATE Generate final synthesized response: $R \leftarrow \text{LLM}_{\text{Main}}(X_{\text{Main}})$
\STATE
\STATE // \textit{Stage 3: State separation and token reduction update}
\STATE Update Main LLM memory: $\mathcal{H}_{\text{Main}} \leftarrow \mathcal{H}_{\text{Main}} \cup \{Q, R\}$
\STATE Update SRM memory: $\mathcal{H}_{\text{SRM}} \leftarrow \mathcal{H}_{\text{SRM}} \cup \{Q, O_{\text{SRM}}\}$ \COMMENT{Excludes $R$ to preserve the SRM's context space}
\STATE
\RETURN $R$
\end{algorithmic}
\end{algorithm*}

\section{Full Results}
\label{sec:full_results}

The following tables present the comprehensive results of our experiments on the Berkeley Function Calling Leaderboard (BFCL). We compare our Instruction-Followed Function Calling (IFFC) framework (in both FP16 and Q4KM quantization) against Prompt Function Calling (PFC) and Native Function Calling (NFC).

\textbf{Notes:}
\begin{itemize}
    \item \textbf{Gemma-3} models do not support NFC; these entries are marked as (-).
    \item \textbf{Granite-4} models were primarily evaluated in Q4KM due to resource constraints.
    \item In the \textit{Live Parallel} and \textit{Live Parallel Multiple} categories, IFFC FP16 results are omitted for most models due to time constraints during the evaluation window.
    \item The IFFC columns are highlighted in gray for clarity.
\end{itemize}

% ==========================================
% TABLE 1: NON-LIVE (SIMPLE & MULTIPLE)
% ==========================================
\begin{table*}[h!]
\centering
\caption{Non-Live Evaluation: Simple and Multiple Categories}
\label{tab:nonlive_simple_multi}
\resizebox{\textwidth}{!}{%
\begin{tabular}{l|cc|cc||cc|cc}
\toprule
 & \multicolumn{4}{c||}{\textbf{Non-Live Simple}} & \multicolumn{4}{c}{\textbf{Non-Live Multiple}} \\
\midrule
\textbf{Model} & \cellcolor{lightgray}\textbf{IFFC FP16} & \cellcolor{lightgray}\textbf{IFFC Q4KM} & \textbf{PFC} & \textbf{NFC} & \cellcolor{lightgray}\textbf{IFFC FP16} & \cellcolor{lightgray}\textbf{IFFC Q4KM} & \textbf{PFC} & \textbf{NFC} \\
\midrule
Gemma-3 1B & \cellcolor{lightgray}21.8\% & \cellcolor{lightgray}3.6\% & 43.5\% & - & \cellcolor{lightgray}36.0\% & \cellcolor{lightgray}21.5\% & 38.5\% & - \\
Gemma-3 4B & \cellcolor{lightgray}87.6\% & \cellcolor{lightgray}87.5\% & 64.3\% & - & \cellcolor{lightgray}85.0\% & \cellcolor{lightgray}84.0\% & 91.5\% & - \\
Gemma-3 12B & \cellcolor{lightgray}93.2\% & \cellcolor{lightgray}94.0\% & 77.3\% & - & \cellcolor{lightgray}93.0\% & \cellcolor{lightgray}93.5\% & 95.0\% & - \\
\midrule
Phi-4 Mini & \cellcolor{lightgray}62.7\% & \cellcolor{lightgray}72.7\% & 67.9\% & 38.0\% & \cellcolor{lightgray}76.5\% & \cellcolor{lightgray}63.5\% & 69.0\% & 0.0\% \\
\midrule
Granite-4 Micro & \cellcolor{lightgray}- & \cellcolor{lightgray}81.2\% & - & - & \cellcolor{lightgray}- & \cellcolor{lightgray}82.0\% & - & - \\
Granite-4 Tiny H & \cellcolor{lightgray}- & \cellcolor{lightgray}46.5\% & - & - & \cellcolor{lightgray}- & \cellcolor{lightgray}31.5\% & - & - \\
\midrule
Qwen-3 0.6B (NoThink) & \cellcolor{lightgray}53.7\% & \cellcolor{lightgray}4.2\% & - & - & \cellcolor{lightgray}68.0\% & \cellcolor{lightgray}19.0\% & - & - \\
Qwen-3 0.6B (Think) & \cellcolor{lightgray}82.0\% & \cellcolor{lightgray}74.5\% & 64.0\% & 62.3\% & \cellcolor{lightgray}89.5\% & \cellcolor{lightgray}76.5\% & 89.0\% & 88.0\% \\
\midrule
Qwen-3 1.7B (NoThink) & \cellcolor{lightgray}87.2\% & \cellcolor{lightgray}84.5\% & - & - & \cellcolor{lightgray}78.5\% & \cellcolor{lightgray}81.5\% & - & - \\
Qwen-3 1.7B (Think) & \cellcolor{lightgray}80.1\% & \cellcolor{lightgray}80.5\% & - & 71.1\% & \cellcolor{lightgray}81.0\% & \cellcolor{lightgray}84.5\% & 92.5\% & 93.0\% \\
\midrule
Qwen-3 4B (NoThink) & \cellcolor{lightgray}94.0\% & \cellcolor{lightgray}93.5\% & - & - & \cellcolor{lightgray}95.5\% & \cellcolor{lightgray}94.0\% & - & - \\
Qwen-3 4B (Think) & \cellcolor{lightgray}96.0\% & \cellcolor{lightgray}95.5\% & 76.1\% & 75.3\% & \cellcolor{lightgray}97.5\% & \cellcolor{lightgray}96.5\% & 97.0\% & 96.5\% \\
\midrule
Qwen-3 8B (NoThink) & \cellcolor{lightgray}96.0\% & \cellcolor{lightgray}95.7\% & - & - & \cellcolor{lightgray}96.0\% & \cellcolor{lightgray}96.0\% & - & - \\
Qwen-3 8B (Think) & \cellcolor{lightgray}96.5\% & \cellcolor{lightgray}95.7\% & 78.4\% & 76.8\% & \cellcolor{lightgray}97.5\% & \cellcolor{lightgray}96.5\% & 96.0\% & 95.5\% \\
\bottomrule
\end{tabular}
}
\end{table*}

% ==========================================
% TABLE 2: NON-LIVE (PARALLEL & PARALLEL MULTIPLE)
% ==========================================
\begin{table*}[h!]
\centering
\caption{Non-Live Evaluation: Parallel and Parallel Multiple Categories}
\label{tab:nonlive_parallel}
\resizebox{\textwidth}{!}{%
\begin{tabular}{l|cc|cc||cc|cc}
\toprule
 & \multicolumn{4}{c||}{\textbf{Non-Live Parallel}} & \multicolumn{4}{c}{\textbf{Non-Live Parallel Multiple}} \\
\midrule
\textbf{Model} & \cellcolor{lightgray}\textbf{IFFC FP16} & \cellcolor{lightgray}\textbf{IFFC Q4KM} & \textbf{PFC} & \textbf{NFC} & \cellcolor{lightgray}\textbf{IFFC FP16} & \cellcolor{lightgray}\textbf{IFFC Q4KM} & \textbf{PFC} & \textbf{NFC} \\
\midrule
Gemma-3 1B & \cellcolor{lightgray}22.0\% & \cellcolor{lightgray}- & 2.0\% & - & \cellcolor{lightgray}16.0\% & \cellcolor{lightgray}9.5\% & 2.0\% & - \\
Gemma-3 4B & \cellcolor{lightgray}82.5\% & \cellcolor{lightgray}- & 56.5\% & - & \cellcolor{lightgray}72.5\% & \cellcolor{lightgray}70.0\% & 41.0\% & - \\
Gemma-3 12B & \cellcolor{lightgray}90.0\% & \cellcolor{lightgray}- & 90.0\% & - & \cellcolor{lightgray}89.0\% & \cellcolor{lightgray}88.5\% & 73.0\% & - \\
\midrule
Phi-4 Mini & \cellcolor{lightgray}78.5\% & \cellcolor{lightgray}- & 16.0\% & 0.0\% & \cellcolor{lightgray}82.0\% & \cellcolor{lightgray}56.5\% & 14.5\% & 0.0\% \\
\midrule
Granite-4 Micro & \cellcolor{lightgray}- & \cellcolor{lightgray}80.0\% & - & - & \cellcolor{lightgray}- & \cellcolor{lightgray}78.5\% & - & - \\
Granite-4 Tiny H & \cellcolor{lightgray}- & \cellcolor{lightgray}70.5\% & - & - & \cellcolor{lightgray}- & \cellcolor{lightgray}71.5\% & - & - \\
\midrule
Qwen-3 0.6B (NoThink) & \cellcolor{lightgray}65.0\% & \cellcolor{lightgray}18.5\% & - & - & \cellcolor{lightgray}62.5\% & \cellcolor{lightgray}25.5\% & - & - \\
Qwen-3 0.6B (Think) & \cellcolor{lightgray}69.0\% & \cellcolor{lightgray}69.5\% & 75.0\% & 69.0\% & \cellcolor{lightgray}65.0\% & \cellcolor{lightgray}56.5\% & 63.0\% & 68.0\% \\
\midrule
Qwen-3 1.7B (NoThink) & \cellcolor{lightgray}85.0\% & \cellcolor{lightgray}77.5\% & - & - & \cellcolor{lightgray}80.5\% & \cellcolor{lightgray}78.5\% & - & - \\
Qwen-3 1.7B (Think) & \cellcolor{lightgray}89.0\% & \cellcolor{lightgray}84.5\% & 88.0\% & 87.5\% & \cellcolor{lightgray}81.0\% & \cellcolor{lightgray}82.0\% & 81.5\% & 81.0\% \\
\midrule
Qwen-3 4B (NoThink) & \cellcolor{lightgray}89.5\% & \cellcolor{lightgray}89.5\% & - & - & \cellcolor{lightgray}88.5\% & \cellcolor{lightgray}87.0\% & - & - \\
Qwen-3 4B (Think) & \cellcolor{lightgray}92.5\% & \cellcolor{lightgray}92.0\% & 92.0\% & 92.0\% & \cellcolor{lightgray}90.5\% & \cellcolor{lightgray}90.0\% & 89.5\% & 90.5\% \\
\midrule
Qwen-3 8B (NoThink) & \cellcolor{lightgray}93.5\% & \cellcolor{lightgray}91.0\% & - & - & \cellcolor{lightgray}89.5\% & \cellcolor{lightgray}88.5\% & - & - \\
Qwen-3 8B (Think) & \cellcolor{lightgray}93.5\% & \cellcolor{lightgray}95.0\% & 95.0\% & 94.5\% & \cellcolor{lightgray}89.5\% & \cellcolor{lightgray}90.5\% & 89.5\% & 88.5\% \\
\bottomrule
\end{tabular}
}
\end{table*}

% ==========================================
% TABLE 3: LIVE (SIMPLE & MULTIPLE)
% ==========================================
\begin{table*}[h!]
\centering
\caption{Live Evaluation: Simple and Multiple Categories}
\label{tab:live_simple_multi}
\resizebox{\textwidth}{!}{%
\begin{tabular}{l|cc|cc||cc|cc}
\toprule
 & \multicolumn{4}{c||}{\textbf{Live Simple}} & \multicolumn{4}{c}{\textbf{Live Multiple}} \\
\midrule
\textbf{Model} & \cellcolor{lightgray}\textbf{IFFC FP16} & \cellcolor{lightgray}\textbf{IFFC Q4KM} & \textbf{PFC} & \textbf{NFC} & \cellcolor{lightgray}\textbf{IFFC FP16} & \cellcolor{lightgray}\textbf{IFFC Q4KM} & \textbf{PFC} & \textbf{NFC} \\
\midrule
Gemma-3 1B & \cellcolor{lightgray}13.9\% & \cellcolor{lightgray}8.8\% & 30.0\% & - & \cellcolor{lightgray}3.1\% & \cellcolor{lightgray}7.1\% & 10.5\% & - \\
Gemma-3 4B & \cellcolor{lightgray}72.4\% & \cellcolor{lightgray}77.1\% & 72.9\% & - & \cellcolor{lightgray}63.5\% & \cellcolor{lightgray}63.7\% & 62.8\% & - \\
Gemma-3 12B & \cellcolor{lightgray}85.2\% & \cellcolor{lightgray}86.4\% & 84.9\% & - & \cellcolor{lightgray}74.5\% & \cellcolor{lightgray}78.5\% & 70.9\% & - \\
\midrule
Phi-4 Mini & \cellcolor{lightgray}42.0\% & \cellcolor{lightgray}39.0\% & 55.0\% & 40.3\% & \cellcolor{lightgray}- & \cellcolor{lightgray}47.6\% & 59.5\% & 50.3\% \\
\midrule
Granite-4 Micro & \cellcolor{lightgray}- & \cellcolor{lightgray}68.9\% & - & - & \cellcolor{lightgray}- & \cellcolor{lightgray}60.4\% & - & - \\
Granite-4 Tiny H & \cellcolor{lightgray}- & \cellcolor{lightgray}31.7\% & - & - & \cellcolor{lightgray}- & \cellcolor{lightgray}35.1\% & - & - \\
\midrule
Qwen-3 0.6B (NoThink) & \cellcolor{lightgray}50.0\% & \cellcolor{lightgray}3.1\% & - & - & \cellcolor{lightgray}29.1\% & \cellcolor{lightgray}2.0\% & - & - \\
Qwen-3 0.6B (Think) & \cellcolor{lightgray}66.2\% & \cellcolor{lightgray}58.5\% & 66.1\% & 65.9\% & \cellcolor{lightgray}52.2\% & \cellcolor{lightgray}47.5\% & 52.2\% & 54.4\% \\
\midrule
Qwen-3 1.7B (NoThink) & \cellcolor{lightgray}66.6\% & \cellcolor{lightgray}58.1\% & - & - & \cellcolor{lightgray}53.1\% & \cellcolor{lightgray}60.2\% & - & - \\
Qwen-3 1.7B (Think) & \cellcolor{lightgray}75.9\% & \cellcolor{lightgray}71.7\% & 75.6\% & 75.6\% & \cellcolor{lightgray}68.6\% & \cellcolor{lightgray}73.5\% & 68.6\% & 72.6\% \\
\midrule
Qwen-3 4B (NoThink) & \cellcolor{lightgray}78.6\% & \cellcolor{lightgray}75.1\% & - & - & \cellcolor{lightgray}73.8\% & \cellcolor{lightgray}73.4\% & - & - \\
Qwen-3 4B (Think) & \cellcolor{lightgray}90.3\% & \cellcolor{lightgray}86.4\% & 87.9\% & 87.6\% & \cellcolor{lightgray}81.5\% & \cellcolor{lightgray}79.9\% & 80.7\% & 79.9\% \\
\midrule
Qwen-3 8B (NoThink) & \cellcolor{lightgray}73.6\% & \cellcolor{lightgray}73.6\% & - & - & \cellcolor{lightgray}78.6\% & \cellcolor{lightgray}76.0\% & - & - \\
Qwen-3 8B (Think) & \cellcolor{lightgray}87.9\% & \cellcolor{lightgray}87.2\% & 87.2\% & 84.9\% & \cellcolor{lightgray}80.4\% & \cellcolor{lightgray}80.7\% & 79.4\% & 79.4\% \\
\bottomrule
\end{tabular}
}
\end{table*}

% ==========================================
% TABLE 4: LIVE (PARALLEL & PARALLEL MULTIPLE)
% ==========================================
\begin{table*}[h!]
\centering
\caption{Live Evaluation: Parallel and Parallel Multiple Categories}
\label{tab:live_parallel}
\resizebox{\textwidth}{!}{%
\begin{tabular}{l|cc|cc||cc|cc}
\toprule
 & \multicolumn{4}{c||}{\textbf{Live Parallel}} & \multicolumn{4}{c}{\textbf{Live Parallel Multiple}} \\
\midrule
\textbf{Model} & \cellcolor{lightgray}\textbf{IFFC FP16} & \cellcolor{lightgray}\textbf{IFFC Q4KM} & \textbf{PFC} & \textbf{NFC} & \cellcolor{lightgray}\textbf{IFFC FP16} & \cellcolor{lightgray}\textbf{IFFC Q4KM} & \textbf{PFC} & \textbf{NFC} \\
\midrule
Gemma-3 1B & \cellcolor{lightgray}31.3\% & \cellcolor{lightgray}0.0\% & 0.0\% & - & \cellcolor{lightgray}8.3\% & \cellcolor{lightgray}0.0\% & 0.0\% & - \\
Gemma-3 4B & \cellcolor{lightgray}68.8\% & \cellcolor{lightgray}75.0\% & 37.5\% & - & \cellcolor{lightgray}54.2\% & \cellcolor{lightgray}49.9\% & 29.2\% & - \\
Gemma-3 12B & \cellcolor{lightgray}87.5\% & \cellcolor{lightgray}87.5\% & 87.5\% & - & \cellcolor{lightgray}79.2\% & \cellcolor{lightgray}79.2\% & 62.5\% & - \\
\midrule
Phi-4 Mini & \cellcolor{lightgray}- & \cellcolor{lightgray}50.0\% & 68.8\% & 0.0\% & \cellcolor{lightgray}- & \cellcolor{lightgray}37.5\% & 66.7\% & 29.2\% \\
\midrule
Granite-4 Micro & \cellcolor{lightgray}- & \cellcolor{lightgray}75.0\% & - & - & \cellcolor{lightgray}- & \cellcolor{lightgray}- & - & - \\
Granite-4 Tiny H & \cellcolor{lightgray}- & \cellcolor{lightgray}18.8\% & - & - & \cellcolor{lightgray}- & \cellcolor{lightgray}- & - & - \\
\midrule
Qwen-3 0.6B (NoThink) & \cellcolor{lightgray}- & \cellcolor{lightgray}0.0\% & - & - & \cellcolor{lightgray}- & \cellcolor{lightgray}12.5\% & - & - \\
Qwen-3 0.6B (Think) & \cellcolor{lightgray}- & \cellcolor{lightgray}62.5\% & 40.3\% & 37.5\% & \cellcolor{lightgray}- & \cellcolor{lightgray}50.0\% & 54.2\% & 54.2\% \\
\midrule
Qwen-3 1.7B (NoThink) & \cellcolor{lightgray}- & \cellcolor{lightgray}62.5\% & - & - & \cellcolor{lightgray}- & \cellcolor{lightgray}70.8\% & - & - \\
Qwen-3 1.7B (Think) & \cellcolor{lightgray}- & \cellcolor{lightgray}75.0\% & 75.0\% & 75.0\% & \cellcolor{lightgray}- & \cellcolor{lightgray}66.6\% & 75.0\% & 75.0\% \\
\midrule
Qwen-3 4B (NoThink) & \cellcolor{lightgray}- & \cellcolor{lightgray}75.0\% & - & - & \cellcolor{lightgray}- & \cellcolor{lightgray}70.8\% & - & - \\
Qwen-3 4B (Think) & \cellcolor{lightgray}- & \cellcolor{lightgray}87.5\% & 77.5\% & 75.0\% & \cellcolor{lightgray}- & \cellcolor{lightgray}87.5\% & 85.0\% & 83.3\% \\
\midrule
Qwen-3 8B (NoThink) & \cellcolor{lightgray}- & \cellcolor{lightgray}75.0\% & - & - & \cellcolor{lightgray}- & \cellcolor{lightgray}70.8\% & - & - \\
Qwen-3 8B (Think) & \cellcolor{lightgray}- & \cellcolor{lightgray}81.3\% & 67.3\% & 62.5\% & \cellcolor{lightgray}- & \cellcolor{lightgray}83.3\% & 79.2\% & 75.0\% \\
\bottomrule
\end{tabular}
}
\end{table*}

% ==========================================
% TABLE 5: PROPRIETARY MODELS
% ==========================================
\begin{table*}[h!]
\centering
\caption{Proprietary Models Baseline (Native Function Calling)}
\label{tab:proprietary_baselines}
\resizebox{\textwidth}{!}{%
\begin{tabular}{l|c c c c c c c | c c}
\toprule
 & \multicolumn{7}{c|}{\textbf{General Purpose Models}} & \multicolumn{2}{c}{\textbf{Claude Series}} \\
\midrule
\textbf{Category} & \textbf{GPT 4.1} & \textbf{GPT 4.1} & \textbf{Gemini} & \textbf{Grok 4} & \textbf{Kimi} & \textbf{O4} & \textbf{GPT} & \textbf{Claude} & \textbf{Claude} \\
 & \textbf{Mini} & & \textbf{2.5 Pro} & \textbf{07-09} & \textbf{K2 Inst.} & \textbf{mini} & \textbf{5.2} & \textbf{Opus 4.5} & \textbf{Sonnet 4.5} \\
\midrule
Simple & 73.8\% & 74.2\% & 66.4\% & 73.5\% & 78.2\% & 70.6\% & 72.9\% & 76.8\% & 72.6\% \\
Multiple & 93.5\% & 90.5\% & 86.0\% & 92.5\% & 93.0\% & 84.5\% & 88.0\% & 95.5\% & 95.5\% \\
Parallel & 93.0\% & 91.0\% & 69.0\% & 88.5\% & 85.5\% & 0.0\% & 89.0\% & 93.5\% & 94.5\% \\
Parallel Multiple & 87.0\% & 86.0\% & 40.0\% & 87.0\% & 84.0\% & 0.0\% & 77.5\% & 88.5\% & 92.0\% \\
\midrule
Live Simple & 80.6\% & 80.2\% & 77.9\% & 82.2\% & 88.0\% & 68.6\% & 71.7\% & 86.4\% & 89.5\% \\
Live Multiple & 77.8\% & 78.4\% & 62.2\% & 73.9\% & 79.4\% & 68.1\% & 70.4\% & 48.2\% & 78.9\% \\
Live Parallel & 75.0\% & 68.8\% & 68.8\% & 75.0\% & 87.5\% & 0.0\% & 68.8\% & 87.5\% & 87.5\% \\
Live Par. Multi. & 66.7\% & 66.7\% & 62.5\% & 79.2\% & 62.5\% & 0.0\% & 58.3\% & 75.0\% & 83.3\% \\
\bottomrule
\end{tabular}
}
\end{table*}

\end{document}